\documentclass[10pt,twocolumn,letterpaper]{article}

\PassOptionsToPackage{table,dvipsnames}{xcolor}

\usepackage[pagenumbers]{cvpr}
\usepackage{stfloats}
\usepackage{graphicx}
\usepackage{pifont}
\usepackage{tcolorbox}
\usepackage{booktabs}
\usepackage{multirow}
\usepackage{tabularx}
\usepackage{algorithm}
\usepackage{algorithmic}
\usepackage{subcaption}

\definecolor{cvprblue}{rgb}{0.21,0.49,0.74}
\usepackage[pagebackref,breaklinks,colorlinks,allcolors=cvprblue]{hyperref}

\def\paperID{4010} 
\def\confName{CVPR}
\def\confYear{2025}

\title{AToM: Aligning Text-to-Motion Model at Event-Level with GPT-4Vision Reward}

\author{
Haonan Han$^{1}$\footnotemark[1]
\and
Xiangzuo Wu$^{1}$\footnotemark[1]
\and
Huan Liao$^{1}$\footnotemark[1]
\and
Zunnan Xu$^{1}$
\and  
Zhongyuan Hu$^{1}$
\and
Ronghui Li$^{1}$
\and 
Yachao Zhang$^{2}$\footnotemark[2]
\and 
Xiu Li$^{1}$\footnotemark[2]
\and
{\small
$^{1}$ Shenzhen International Graduate School, Tsinghua University \quad
$^{2}$ School of Informatics, Xiamen University
}
}

\begin{document}
\twocolumn[{%
\renewcommand\twocolumn[1][]{#1}%
\maketitle


\begin{center}
    \centering
    \captionsetup{type=figure}     
    \vspace{-1em}
    \includegraphics[width=\textwidth]{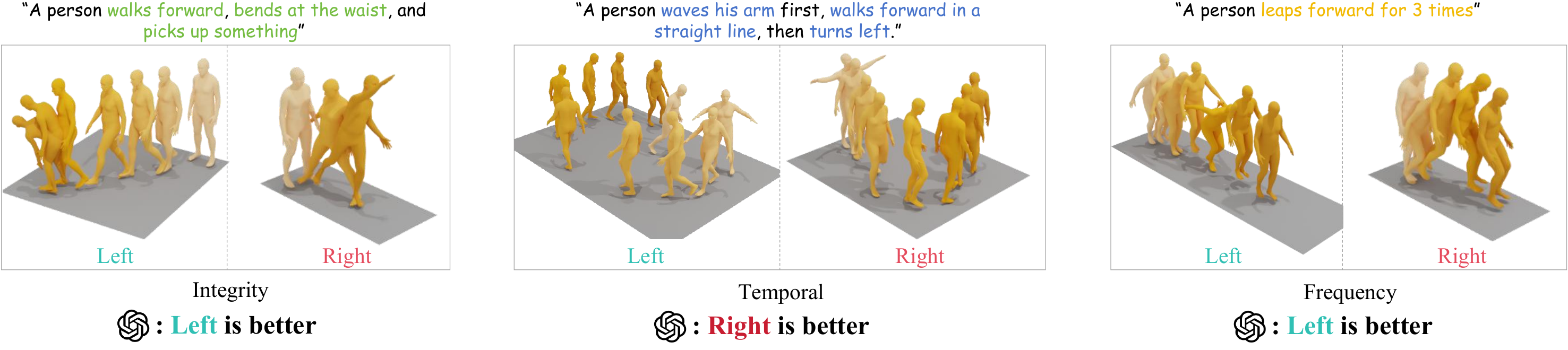}
     \vspace{-2em}
    \captionof{figure}{Showcases of motion samples for three scenarios. The two motion samples for each scenario were generated based on the prompt above the samples. Moreover, we leverage GPT-4V to compare two motion samples according to the degree of alignment between the motion samples and the input prompt.}
    \label{fig:motivation}
\end{center}%
}]

\begin{abstract}
\let\thefootnote\relax\footnotetext{\noindent{$^*$ Equal contribution. $^\dagger$ Corresponding authors.}}
Recently, text-to-motion models have opened new possibilities for creating realistic human motion with greater efficiency and flexibility. However, aligning motion generation with event-level textual descriptions presents unique challenges due to the complex relationship between textual prompts and desired motion outcomes. To address this, we introduce \textbf{AToM}, a framework that enhances the alignment between generated motion and text prompts by leveraging reward from GPT-4Vision. AToM comprises three main stages: Firstly, we construct a dataset \textnormal{\texttt{MotionPrefer}} that pairs three types of event-level textual prompts with generated motions, which cover the integrity, temporal relationship and frequency of motion. Secondly, we design a paradigm that utilizes GPT-4Vision for detailed motion annotation, including visual data formatting, task-specific instructions and scoring rules for each sub-task. Finally, we fine-tune an existing text-to-motion model using reinforcement learning guided by this paradigm. Experimental results demonstrate that AToM significantly improves the event-level alignment quality of text-to-motion generation. Project page is available at \href{https://atom-motion.github.io/}{https://atom-motion.github.io/}. 
\end{abstract}
\vspace{-2em}    
\section{Introduction}
Generating high-quality human motions from textual descriptions is a promising task and plays an important role in fields such as game production, animation, film and virtual reality. Recently, generative models leveraging autoregressive~\cite{jiang2023motiongpt, zhang2023generating, guo2022tm2t, guo2024momask} and diffusion-based approaches~\cite{Guy2022mdm, chen2023executing, zhang2023remodiffuse, shafir2023human} have shown remarkable performance on this task. These models generally perform well with short text prompts. 
However, due to the scarcity of text-motion pairs as well as the coarse-grained text descriptions that cover limited motion scenarios, there are challenges in mapping complex descriptions (e.g., multi-motion events or motions with temporal relationships and specified frequency) to corresponding motion sequences, limiting the model’s ability to generalize effectively.
As shown in the second group of Figure~\ref{fig:motivation}, given a prompt describing three motions in temporal order, the left example motion accurately captures the expression ``waves his arm" but misinterprets ``walks forward in a straight line" as ``walks in a circle" and fails to account for the ``turning left" action.

One potential solution to address this issue is to alleviate data scarcity by collecting additional text-motion pair data. However, unlike tasks involving language or image data, gathering motion data typically requires specialized motion capture equipment and expert annotations, which are both costly and labor-intensive. An alternative approach is to leverage the success of language models by fine-tuning pre-trained models with preference data, thereby enhancing their alignment capabilities. To the best of our knowledge, InstructMotion~\cite{sheng2024exploring} is the first work to fine-tune a text-to-motion model using human preference data through reinforcement learning from human feedback (RLHF), leveraging human-labeled data to enhance model alignment with preferred outputs. Subsequent approaches~\cite{mao2024learning, pappa2024modipo} have explored the use of automated preference datasets, employing pretrained models and reward functions to approximate human feedback. Such methods aim to reduce reliance on manual annotation and improve model performance across a range of alignment metrics

While human annotation reduces workload, it remains labor-intensive and challenging to scale. AI feedback-based methods reduce reliance on human annotators but rely on models trained on standard motion datasets, like HumanML3D~\cite{Guo_2022_CVPR}, limiting their capacity to score out-of-distribution text-motion pairs. Additionally, both human and AI-based approaches often treat text and motion data as unified wholes, overlooking the need for fine-grained alignment evaluation, such as event-level correspondence.

To address these limitations, we propose leveraging advancements in rapidly evolving Vision-Language Large Models, such as GPT-4Vision. These models, through sophisticated architectures and large-scale training, have demonstrated state-of-the-art performance in tasks requiring seamless integration of textual and visual information, making them well-suited for the text-motion alignment task. By utilizing GPT-4Vision, we aim to simultaneously address challenges such as data scarcity, labor-intensive annotation, scalability issues, and the need for granular alignment evaluation—thus paving the way for more robust, scalable, and accurate alignment solutions.

To investigate GPT-4Vision's ability in aligning text and motion modalities, we propose a new framework named AToM, designed for aligning text-to-motion models using feedback from GPT-4Vision or other Vision-Language Large Models.  AToM consists of three main stages: \textbf{(1)} we generate initial text prompts using GPT-4. These prompts are then fed into a motion generation model to generate several different motions for each text prompt. For ease of comparison with human-based work InstructMotion~\cite{sheng2024exploring}, we also adopt MotionGPT~\cite{jiang2023motiongpt} as our motion generator. \textbf{(2)} We then render these generated motions and sample the rendered motion as a sequence of sampled frames. 
\begin{table}[t]
    \centering
    \caption{Statistics of existing preference datasets for text-to-motion generative models. “Fine Grained” represents containing preference regarding multiple aspects or not.}
    \label{tab:preference_datasets}
    \resizebox{\columnwidth}{!}{ 
    \begin{tabular}{l|l|ccc}
        \toprule
        \textbf{\small{Dataset}} & \textbf{\small{Annotator}} & \textbf{\small{Prompts}} & \textbf{\small{Pairs}} & \textbf{\small{Fine Grained}} \\
        \midrule
        InstructMotion~\cite{sheng2024exploring} & Human & 3.5K & 3.5K & \textcolor{red}{\ding{55}} \\
        \midrule
        \multirow{2}{*}{Pick-a-Move~\cite{pappa2024modipo}} & TMR~\cite{Petrovich_2023_ICCV} & -- & -- & \textcolor{red}{\ding{55}} \\
        & Guo \etal~\cite{Guo_2022_CVPR} & -- & -- & \textcolor{red}{\ding{55}} \\
        \midrule
        \small{\texttt{MotionPrefer} (ours)} & GPT-4V~\cite{openai2024gpt4technicalreport} & \textbf{5.3K} & \textbf{80K} & \textcolor{ForestGreen}{\checkmark} \\
        \bottomrule
    \end{tabular}
    }
    \vspace{-0.4cm}
\end{table}
The input text prompt, sampled frames, along with an instruction describing the alignment rules, are then fed into the GPT-4Vision model. And the model will evaluate the text-motion alignment score at event-level based on the given frames and rules. The text prompts, generated motions, and the corresponding alignment scores jointly constitute our \texttt{MotionPrefer} dataset. In Table~\ref{tab:preference_datasets}, we compare our \texttt{MotionPrefer} dataset with two other preference datasets, InstructMotion~\cite{sheng2024exploring} and Pick-a-Move~\cite{pappa2024modipo}. In terms of the scale, our \texttt{MotionPrefer} dataset surpasses InstructMotion~\cite{sheng2024exploring}, while the Pick-a-Move~\cite{pappa2024modipo} does not specify its data volume. Furthermore, our dataset is the only one to provide fine-grained preference annotations across multiple aspects, including motion integrity, temporal order, and frequency. \textbf{(3)} We finetune the motion generator, MotionGPT~\cite{jiang2023motiongpt}, on our \texttt{MotionPrefer} dataset with LoRA\cite{hu2021lora} and IPO\cite{azar2024general} RL strategy.

We summarize our contributions as follows:

\begin{itemize}

    \item We created a dataset named \texttt{MotionPrefer}, consisting of 5.3K text prompts and 80K motion preference pairs. Each text-motion pair is scored based on event-level correspondences which contain three dimensions: motion integrity, temporal order, and motion frequency, thereby surpassing existing datasets in both scale and quality.
    \item We designed an annotation and reward paradigm leveraging GPT-4V, which encompasses key elements such as motion instructions, motion injection methods and scoring rules for each sub-task. This comprehensive paradigm is applied to evaluate the text-motion pairs collected in the \texttt{MotionPrefer} dataset, providing preference-based reward scoring.
    \item We used AToM to fine-tune an off-the-shelf motion generator across three sub-tasks: motion integrity, temporal order and frequency, achieving substantial performance improvements. Additionally, ablation studies explored the effects of various motion injection methods, LoRA, score filtering, and reinforcement learning strategies. The experimental results strongly support AToM's effectiveness in improving text-motion alignment across tasks.
    
\end{itemize}

\section{Related Works}
\subsection{Text-to-Motion Generative Models}
Text-to-motion (T2M) generation aims to generate human motion that corresponds to free-form natural language descriptions, serving as a fundamental task in the field of motion generation. Text2Action \cite{ahn2018text2action} pioneered this area by utilizing a GAN based on a SEQ2SEQ model to map short descriptions to human actions. Language2Pose \cite{ahuja2019language2pose} introduced a curriculum learning approach to develop joint-level embeddings for text and pose, while Lin et al. \cite{lin2018generating} proposed an end-to-end SEQ2SEQ model for generating more realistic animations. However, the limited availability of large-scale supervised datasets hinders generalization to novel descriptions, such as unseen combinations of motions.

To address these issues, Ghosh et al. \cite{ghosh2021synthesis} developed a hierarchical two-stream sequential model capable of handling long sentences that describe multiple actions. MotionCLIP \cite{tevet2022motionclip} aligns the human motion manifold with CLIP space to endow the model with zero-shot capabilities.
More recent work, such as the Transformer-based TEACH \cite{athanasiou2022teach}, generates realistic 3D human motions that follow complex, sequential action instructions, facilitating flexible temporal action composition. TEMOS \cite{zhang2024temo} uses a Transformer-based VAE and an additional text encoder for multi-object 3D scene generation and editing, guided by multi-level contrastive supervision. T2M-GPT \cite{zhang2023generating} combines VQ-VAE and GPT to obtain high-quality discrete representations, achieving competitive motion generation results. 
The diffusion-based model MotionDiffuse \cite{zhang2022motiondiffuse} allows for fine-grained control over body parts and supports arbitrary-length sequences through a series of denoising steps with injected variations. The classifier-free diffusion model MDM \cite{Guy2022mdm} predicts motion samples instead of noise, facilitating geometric loss application and setting state-of-the-art performance. MLD \cite{chen2023executing} further advances motion generation using a latent diffusion model. MotionGPT\cite{jiang2023motiongpt} develops unified large motion-language models that represent human motion via discrete vector quantization, enabling versatile performance across tasks like motion generation, captioning, and prediction. However, due to the coarse grained motion-paired text descriptions, achieving robust performance in zero-shot and multi-event scenarios remains challenging.

\subsection{Aligning Models with Human/AI Feedback}
Reinforcement Learning from Human Feedback (RLHF) \cite{ouyang2022training, bai2022training} has emerged as a transformative technique for model alignment, especially in applications with complex or ambiguous objectives. It has become the primary approach for aligning large language models (LLMs) \cite{ziegler2019fine, achiam2023gpt, anthropic2024claude} with user intent and has been effectively extended to generative models in image \cite{lee2023aligning, xu2024imagereward} and audio generation \cite{liao2024baton}. In addition to traditional methods like PPO \cite{schulman2017proximal} and RLHF-PPO \cite{ouyang2022training}, which use explicit reward models, alternative approaches such as Direct Preference Optimization (DPO) \cite{rafailov2024direct} and Slic-hf \cite{zhao2023slic} streamline the alignment process. These methods directly optimize model policies based on human preferences, which reduces computational overhead and enables potentially more robust optimization by working directly with preference data. 

In aligning generated motion with human perception, existing methods often struggle with overfitting specific motion expressions due to limited training data, which relies heavily on expert-labeled motion. InstructMotion \cite{sheng2024exploring} addresses this by incorporating human preference data, where non-expert labelers compare generated motions, introducing preference learning to T2M generation and achieving performance improvements over traditional methods.

Given the high cost of obtaining quality preference labels, Reinforcement Learning from AI Feedback (RLAIF) \cite{bai2022constitutional} presents another promising alternative. Mao et al. \cite{mao2024learning} decompose motion descriptions into meta motions, combining them to generate novel descriptions. They use a trial-and-error approach in reinforcement learning, designing a reward model based on contrastive pre-trained text and motion encoders, enhancing semantic alignment and leveraging synthetic text-only data for better generalization. MoDiPO \cite{pappa2024modipo} applies DPO with AI feedback, to confine diffusion-based motion generation within realistic and text-aligned boundaries. Unlike InstructMotion, MoDiPO focuses on optimizing model alignment rather than generalizable generation, ensuring high-quality, contextually consistent outputs.

Our approach significantly diverges from previous work in two
key aspects: 1) Instead of relying on labor-intensive, human-feedback-driven methods, we are the first to utilize a more efficient, LLM-based AI feedback mechanism from GPT-4V. 2)  We are the first to focus on fine-grained alignment between motion and text, particularly at the event level. We emphasize integrity, temporal and frequency correspondence between the motion described in the text and the motion generated, providing a more detailed approach to motion-text alignment.
\section{Method}
\begin{figure*}[t!]
    \centering
    \vspace{-0.6cm}
    \includegraphics[width=\textwidth]{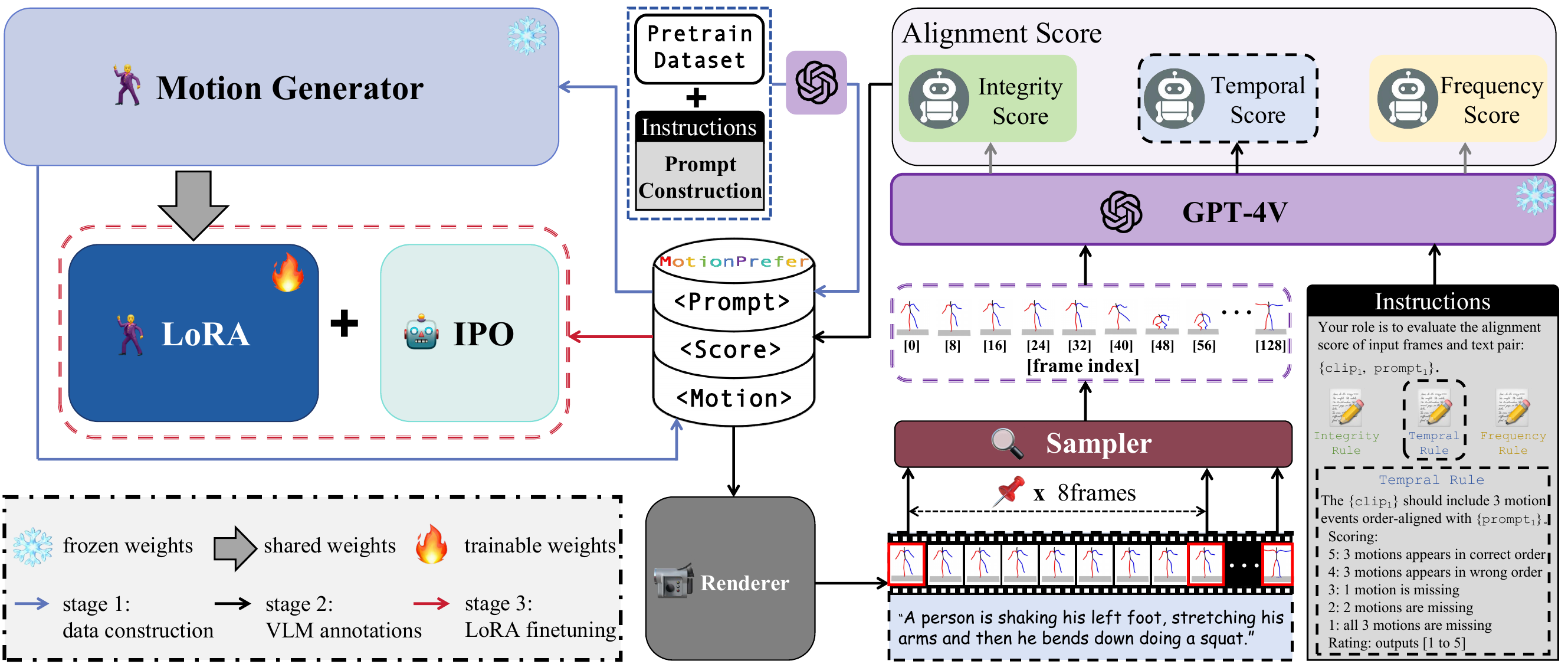}
    \caption{\textbf{The framework of AToM.} AToM encompasses three stages: (1) A motion generation process using task-specific prompts constructed by LLM; (2) Evaluation of alignment score for text-motion pairs using a predefined reward paradigm based on LVLM; (3) A fine-tuning mechanism based on LoRA and RL strategy that enhances the original motion generator using the dataset \small{\texttt{MotionPrefer}}.}
    \label{fig:pipeline}
\end{figure*}
The framework of AToM, as shown in Figure~\ref{fig:pipeline}, consists of three stages. Firstly, we constructed a synthetic dataset of motion-text pairs using task-specific selected prompts and corresponding multiple output motion sequences from the motion generator, illustrated in Section~\ref{dataset}. Furthermore, in Section~\ref{RPD}, We developed a reward paradigm based on GPT-4Vision to score the alignment of visual signals rendered from motion sequences across three aspects. Finally, as elaborated in Section~\ref{FT}, we utilized the synthetic dataset collected in stage 1 and the alignment score annotated in stage 2 to fine-tune the text-to-motion model, thereby enhancing its task-specific alignment performance.
\subsection{Dataset Construction}
\label{dataset}
Prior studies~\cite{sheng2024exploring,pappa2024modipo} have found that text-to-motion models face challenges in producing motions aligning with input textual descriptions, primarily manifest in low integrity, incorrect temporal relationships and frequency. As shown in Figure~\ref{fig:motivation}, the left example exhibits poor integrity because of a missed motion event and the middle example has an incorrect temporal relationship due to a wrong time order, while the right example shows the wrong frequency. Specifically, we focus on evaluating the integrity, temporal relationships, and frequency of the generated motion to assess the effectiveness of AI feedback in text-to-motion model. In stage 1, we targetedly construct prompts and generate text-motion samples related to the above three aspects to facilitate reward evaluation in the later stage.
\begin{table}[t]
\centering
\small
\begin{tcolorbox}[]

Given a label group, $\mathrm{X_{task}}$, the three labels in it will be described into a motion event group:\\ $\{ \texttt{Event}_{1}, \texttt{Event}_{2}, \texttt{Event}_{3} \}$.\\ Then, please join the motion event group with conjunction to form a motion prompt, defined as\\ ``$\texttt{Event}_{1}, \texttt{Conjunction}_{1}, \texttt{Event}_{2}, \texttt{Conjunction}_{2}, \\
\texttt{Event}_{3}$". 
\\
\\
Conjunction list: $\mathrm{X_{conj}}$
\\
\\
Please ensure to randomly select conjunctions from the list and avoid relying on a single conjunction. Please try to generate complete prompts that match human language expressions as much as possible.
\end{tcolorbox}
\caption{GPT instruction for prompt construction in temporal task.}
\vspace{-0.4cm}
\label{tab:gpt_construction}
\end{table}
As the process shown by the blue arrow in Figure~\ref{fig:pipeline}, we randomly select $n$=3.5K motion events from the dataset HumanML3D~\cite{Guo_2022_CVPR} as our meta motion labels which can be denoted as $\mathrm{X_{meta}}=\{l_1, ..., l_{n}\}$. For different sub-tasks, we randomly pick labels from the $\mathrm{X_{meta}}$ to form the task-specified label group $\mathrm{X_{task}}$. Based on the label group $\mathrm{X_{task}}$ and a predefined conjunction group $\mathrm{X_{conj}}$, we instruct GPT-4, named $\mathrm{M_{Language}}$, to generate meaningful and complete sentences matching human language with instruction $\mathrm{I_{prompt}}$:
\begin{equation}
    \mathbb{D}_{\text{prompt}} \sim \mathrm{M_{Language}}(\mathrm{I_{prompt}}|\mathrm{X_{task}}, \mathrm{X_{conj}}),
\end{equation}
As shown in Table~\ref{tab:gpt_construction}, here we give a template of the instruction $\mathrm{I_{prompt}}$ for the temporal sub-task to construct prompt $\mathbb{D}_{\text{prompt}}$. The conjunction list $\mathrm{X_{conj}}$ for temporal sub-task encompasses terms such as ``and”, ``then”, ``followed by”, and so forth. Similarly, but distinctively, for the integrity sub-task, we construct prompts by selecting 2-5 motion events from $\mathrm{X_{task}}$ combined with conjunctions to cover most cases; for the frequency sub-task, prompts are typically formed by selecting a single motion event paired with a frequency-descriptive conjunction (refer to the Appendix for detailed instructions regarding the other two sub-tasks).
Based on the prompt data $\mathbb{D}_{\text{prompt}}$ constructed earlier, we use the widely adopted text-to-motion model, MotionGPT~\cite{jiang2023motiongpt}, named $\mathrm{M_{Motion}}$, as our motion generator to produce motion samples, ranging from 6 to 10 per prompt:
\begin{equation}
    \mathbb{D}_{\text{motion}} \sim \mathrm{M_{Motion}}(\mathbb{D}_{\text{prompt}}).
\end{equation}

The details of \texttt{MotionPrefer} are provided in Table~\ref{tab:motionprefer_count}. In total, the \texttt{MotionPrefer} dataset consists of 5,276 prompts and 47.1k motion samples.

\begin{table}[t]
\centering
\resizebox{\columnwidth}{!}{
    \begin{tabular}{ccccc}
    \toprule
    Amount     & Prompt & Motion & Pair& Motion events\\ 
    \midrule
    Integrity        & 2500  & 25K        & 35.3K&2-5    \\
    Temporal         & 1360  & 13.6K       & 35.3K   &3    \\
    Frequency         & 1416  & 8.5K        & 9.4K    &1  \\
    \midrule
    Total            & 5276  & 47.1K        & 80K   & ---\\
    \bottomrule
    \end{tabular}
}
\caption{Details of amounts of \texttt{MotionPrefer} dataset.}
\label{tab:motionprefer_count}
\vspace{-0.5cm}
\end{table}



\subsection{Reward Paradigm Design}
\label{RPD}
\begin{table*}[h!]
\centering
\small
\vspace{-0.6cm}
\caption{Scoring rules for sub-tasks.}
\vspace{-0.1cm}
\label{tab:annotation}
\begin{tabularx}{\textwidth}{l|X|X}
\toprule
Task & Example & Scoring Criteria \\
\midrule
Integrity & A man walks forward, walks backward, and squats. & 5: All described motions appear in the frames. \\
 &  & 0: Some motions are missing or incomplete in frames. \\
\midrule
Temporal & A person walks forwards doing ballet, then raising one leg, then skipping and raising the other leg. & 5: Three motions appear in correct order.

4: Three motions appear in wrong order.\\
 &  & 3: One motion is missing. \\
 &  & 2: Two motions are missing. \\
 &  & 1: All three motions are missing. \\
\midrule
Frequency & A person jumps forward one time. & 3: The motion is correct and the frequency is accurate. \\
 &  & 2: The motion is present but the frequency is incorrect. \\
 &  & 1: The motion is incorrect, regardless of the frequency. \\
\bottomrule
\end{tabularx}
\vspace{-0.3cm}
\end{table*}

The motion sequences generated by MotionGPT \cite{jiang2023motiongpt} were rendered into video using the renderer. This rendered video sequence is denoted as \( \mathbb{D}_{\text{motion-video}} \):
\begin{equation}
    \mathbb{D}_{\text{motion-video}} \sim \mathrm{R_{\text{motion}}}(\mathbb{D}_{\text{motion}}),
\end{equation}
where \( \mathbb{D}_{\text{motion}} \) represents the original motion data and $\mathrm{R_{\text{motion}}}$ denotes the renderer for obtaining the motion video.

The video is then sampled at 8-frame intervals, extracting a sequence of frame \( F \), using the sampling function \( \mathrm{Sampler} \) as follows:
\begin{equation}
    F = \left\{ f_0, f_8, \dots \right\} = \mathrm{Sampler}(\mathbb{D}_{\text{motion-video}}, 8),
\end{equation}
where \( \mathrm{Sampler} \) selects frames at 8-frame intervals from the rendered video. These sampled motion frames \( F \), along with the corresponding text description \( \mathbb{D}_{\text{prompt}} \), are sequentially injected into GPT-4V for evaluation.

We leverage GPT-4V, named $\mathrm{M_{VL}}$, to assess the alignment between generated motion sequences and textual descriptions. With the specific instructions $\mathrm{I_{score}}$ (refer to the
Appendix for the detailed scoring instructions for the three tasks), GPT-4V evaluates the alignment score across three tasks, based on the scoring criteria \( \mathrm{C_{score}} \) detailed in Table \ref{tab:annotation}. 
This scoring process can be represented as:
\begin{equation}
    \mathbb{D}_{\text{reward}} = \mathrm{M_{VL}}\left((F, \mathbb{D}_{\text{prompt}}), \mathrm{I_{score}}\right),
\end{equation}
where the motion frames and text description are input into GPT-4V, along with instructions $\mathrm{I_{score}}$, to compute the alignment score.

For each task, GPT-4V assigns a score reflecting how well the motion sequence aligns with the text description, considering factors such as the integrity, temporal order, and frequency of motions. Through this process, we obtain a set of text-motion pairs with corresponding scores, denoted as $\mathbb{D}_{\text{motionprefer}}$, which can be represented as:

\begin{equation}
    \mathbb{D}_{\text{motionprefer}} = \left\{ \mathbb{D}_{\text{motion}}, \mathbb{D}_{\text{prompt}}, \mathbb{D}_{\text{reward}} \right\},
\end{equation}
where \( \mathbb{D}_{\text{motion}} \) represents the generated motion sequence, \( \mathbb{D}_{\text{prompt}} \) represents the text description, and the \( \mathbb{D}_{\text{score}} \) is the text-motion alignment score annotated by GPT-4V. This annotation framework enables an evaluation of the generated motions across multiple dimensions, serving as a foundation for the next-step finetuning of the pretrain model.

\subsection{Text-to-Motion Model Fine-tuning}

To fine-tune our motion generator with GPT-4V feedback, we apply algorithm \ref{alg:data_construction} as a method to construct our training data $\mathbb{D}$ from $\mathbb{D}_\text{motionprefer}$, where $\mathbb{D_\text{motionprefer}}$ contains pairs in the form $(m, p, r)$, with $m$, $p$ and $r$ derived from the sets $\mathbb{D_\text{motion}}$, $\mathbb{D_\text{prompt}}$ and $\mathbb{D_\text{reward}}$, respectively. This algorithm firstly filter out motion samples relevant to a specific sub-task. It then groups the samples by identical prompts and ranks each group by their quality scores in descending order. For each prompt group, the algorithm iterates over all possible pairs of motion samples, selecting pairs where the difference in quality scores exceeds a predefined threshold $\delta$. These selected pairs, consisting of a high-quality sample $m_{w}$, a low-quality sample $m_{l}$, and their associated prompt $p$, are then added to the training dataset $\mathbb{D}$. 

\begin{algorithm}
\caption{Paired Data Construction for a Subtask}
\label{alg:data_construction}
\begin{algorithmic}
    \REQUIRE Dataset $\mathbb{D}_{\text{motionprefer}}$
    \ENSURE Filtered Training Set $\mathbb{D}$
    
    \STATE Extract subset $\mathbb{D}_{\text{subtask}} = \{(m, p, r) \in \mathbb{D}_{\text{motionprefer}} \mid \text{issubtask}(p)\}$ 
    \STATE Group $\mathbb{D}_{\text{subtask}}$ by $p$ to form $\mathbb{D}_{\text{grouped}}$
    \STATE Initialize an empty set $\mathbb{D}$
    \FOR{each group $g$ in $\mathbb{D}_{\text{grouped}}$}
        \STATE Sort $g$ in descending order by $r$
        \FOR{each pair $(i, j)$ in $g$ where $i < j$}
            \IF{$r_i - r_j > \delta$}
                \STATE Add $(m_i, m_j, p)$ to $\mathbb{D}$
            \ENDIF
        \ENDFOR
    \ENDFOR
    \RETURN $\mathbb{D}$
\end{algorithmic}
\end{algorithm}

\begin{table*}[htbp]
    \small
    \centering
    \vspace{-0.6cm}
    \caption{\textbf{Comparison of AToM with baselines in different tasks}. AToM\raisebox{0.5ex}{\footnotesize$\spadesuit$} represents the process of mixing preference data from three tasks and randomly selecting a subset of preference data (approximately 3.5K pairs) that matches the size of the RLHF framework InstructMotion, ensuring fair comparison with the baseline model. }
    \label{tab:main_results}
    \begin{tabularx}{\textwidth}{l|l|XXXX|XXX}
        \toprule
        Task & Method & \small{MM Dist$\downarrow$} & \small{Top-1$\uparrow$} & \small{Top-2$\uparrow$} & \small{Top-3$\uparrow$} & \small{FID$\downarrow$} & \small{Diversity$\uparrow$} & \small{MModality$\uparrow$} \\
        \midrule
        \multirow{2}{*}{Temporal} & MotionGPT~\cite{jiang2023motiongpt} & $5.652^{\pm .048}$ & $0.193^{\pm .004}$ & $0.312^{\pm .007}$ & $0.403^{\pm .009}$ & $0.655^{\pm .049}$ & $\mathbf{8.936^{\pm .048}}$ & $\mathbf{3.846^{\pm .048}}$ \\
        & \cellcolor{gray!20}AToM (ours) & \cellcolor{gray!20}$\mathbf{5.576^{\pm .027}}$ & \cellcolor{gray!20}$\mathbf{0.199^{\pm .005}}$ & \cellcolor{gray!20}$\mathbf{0.322^{\pm .006}}$ &\cellcolor{gray!20} $\mathbf{0.412^{\pm .008}}$ & \cellcolor{gray!20}$\mathbf{0.613^{\pm .044}}$ & \cellcolor{gray!20}$8.926^{\pm .086}$ & \cellcolor{gray!20}$3.495^{\pm .147}$ \\
        \midrule
        \multirow{2}{*}{Frequency} & MotionGPT~\cite{jiang2023motiongpt} & $5.328^{\pm .063}$ & $0.182^{\pm .011}$ & $\mathbf{0.315^{\pm .012}}$ & $0.406^{\pm .012}$ & $1.409^{\pm .148}$ & $8.857^{\pm .163}$ & $\mathbf{3.591^{\pm .132}}$ \\
        & \cellcolor{gray!20}AToM (ours) & \cellcolor{gray!20}$\mathbf{5.259^{\pm .045}}$ & \cellcolor{gray!20}$\mathbf{0.183^{\pm .009}}$ & \cellcolor{gray!20}$0.314^{\pm .012}$ & \cellcolor{gray!20}$\mathbf{0.407^{\pm .009}}$ & \cellcolor{gray!20}$\mathbf{1.165^{\pm .124}}$ & \cellcolor{gray!20}$\mathbf{8.861^{\pm .163}}$ & \cellcolor{gray!20}$3.261^{\pm .115}$ \\
        \midrule
        \multirow{2}{*}{Integrity} & MotionGPT~\cite{jiang2023motiongpt} & $5.897^{\pm .036}$ & $0.160^{\pm .007}$ & $0.271^{\pm .007}$ & $0.350^{\pm .006}$ & $\mathbf{0.340^{\pm .023}}$ & $8.815^{\pm .130}$ & $\mathbf{3.752^{\pm .162}}$ \\
        & \cellcolor{gray!20}AToM (ours) & \cellcolor{gray!20}$\mathbf{5.871^{\pm .034}}$ & \cellcolor{gray!20}$\mathbf{0.160^{\pm .007}}$ & \cellcolor{gray!20}$\mathbf{0.274^{\pm .008}}$ & \cellcolor{gray!20}$\mathbf{0.360^{\pm .008}}$ & \cellcolor{gray!20}$0.400^{\pm .040}$ & \cellcolor{gray!20}$\mathbf{8.821^{\pm .180}}$ & \cellcolor{gray!20}$3.511^{\pm .172}$ \\
        \midrule
        \multirow{3}{*}{General} & MotionGPT~\cite{jiang2023motiongpt} & $3.998^{\pm .016}$ & $0.407^{\pm .002}$ & $0.571^{\pm .003}$ & $0.661^{\pm .002}$ & $0.188^{{\pm .006}}$ & $\mathbf{9.452^{\pm .080}}$ & $\mathbf{3.421^{\pm .170}}$ \\
        & InstructMotion~\cite{sheng2024exploring} & $3.947^{\pm .013}$ & $0.402^{\pm .003}$ & $0.570^{\pm .003}$ & $0.663^{\pm .003}$ & $0.260^{\pm .009}$ & $9.411^{\pm .120}$ & $2.534^{\pm .152}$ \\
        & \cellcolor{gray!20}AToM\raisebox{0.5ex}{\footnotesize$\spadesuit$} (ours) & \cellcolor{gray!20}$\mathbf{3.943^{\pm .012}}$ & \cellcolor{gray!20}$\mathbf{0.410^{\pm .008}}$ & \cellcolor{gray!20}$\mathbf{0.575^{\pm .003}}$ & \cellcolor{gray!20}$\mathbf{0.669^{\pm .002}}$ & \cellcolor{gray!20}$\mathbf{0.177^{\pm .008}}$ & \cellcolor{gray!20}$9.401^{\pm .107}$ & \cellcolor{gray!20}$3.025^{\pm .174}$ \\
        
        \bottomrule
    \end{tabularx}
    \vspace{-0.3cm}
\end{table*}
We then follow the definition in IPO\cite{azar2024general} to define IPO loss $h_{\pi}(m_w, m_l, p)$:
\begin{equation}
    h_{\pi}(m_w, m_l, p) = 
        \log ( \frac{\pi (m_w|p) \pi_{ref}(m_l | p) } {\pi (m_l|p) \pi_{ref} (m_w | p)} ),
\end{equation}
where $\pi$ represents our training policy, and $\pi_{ref}$ is the reference policy. The loss IPO optimizes upon is given by:
\begin{equation}
    \mathbb{E}_{(m_w, m_l, p) \sim D} (h_{\pi} (m_w, m_l, p) - \frac{1}{2 \beta} )^2.
\end{equation}

To enhance the adaptability and efficiency of our motion generator during fine-tuning, we employed LoRA\cite{hu2021lora}, which enables us to adjust the model’s parameters with significantly fewer computational resources compared to traditional fine-tuning methods. This fine-tuning approach generalizes well across various sub-tasks. As a result, the motion generator showed marked improvements in output quality, achieving stronger alignment between desired text prompts and the generated motions.

\label{FT}







\section{Experiments}

\subsection{Implementation Details}

\noindent\textbf{Dataset} To evaluate the effectiveness of our AI feedback-driven fine-tuning framework, we initialized from the pretrained MotionGPT checkpoint\footnote{\href{https://huggingface.co/OpenMotionLab/MotionGPT-base}{https://huggingface.co/OpenMotionLab/MotionGPT-base}} and fine-tuned it on three subsets of our preference dataset, MotionPrefer, each including 35k (for temporal), 35k (for integrity) and 9.4k (for frequency) motion preference pairs. 

\noindent\textbf{Implementation Specifics} The optimal hyperparameter configuration included a learning rate of 1e-3, batch size of 32, and 20 epochs, using AdamW as the optimizer. A cosine learning rate scheduler was applied, and PEFT\cite{mangrulkar2022peft} parameters were tuned with LoRA\cite{hu2021lora} (R = 8, $\alpha$ = 16, dropout = 0.05), allowing the model to leverage the parameter-efficient fine-tuning structure. This setup required approximately 12 GB of memory on a single RTX 4090 GPU, ensuring efficient fine-tuning throughout the process.

\noindent\textbf{Evaluation Metrics} For evaluation, we filtered the HumanML3D~\cite{Guo_2022_CVPR} test set, obtaining 418, 506, and 234 text-motion pairs for integrity, temporal, and frequency tasks. Consistent with prior research~\cite{zhang2023generating,sheng2024exploring,wang2024motiongpt}, we focus on motion quality and text-motion alignment. Multi-modal Distance (MM-Dist) calculates the average Euclidean distance between text and motion features, while R-Precision measures motion-to-text retrieval accuracy with Top-1, Top-2, and Top-3 scores based on the model’s ability to rank ground-truth descriptions correctly. Motion quality is assessed using FID to compute distribution distances between generated and real motion features. Diversity measures the average Euclidean distance between randomly sampled motion pairs, and MModality evaluates the variation among motions generated from the same text by averaging distances across 20 sequences per description. For subjective analysis, 50 participants evaluated text-motion alignment on temporal, frequency, and integrity aspects, selecting their preferred case or indicating similarity for each comparison.

\subsection{Main Results}
\begin{figure*}[h]
    \vspace{-0.4cm}
    \centering
    \includegraphics[width=\textwidth]{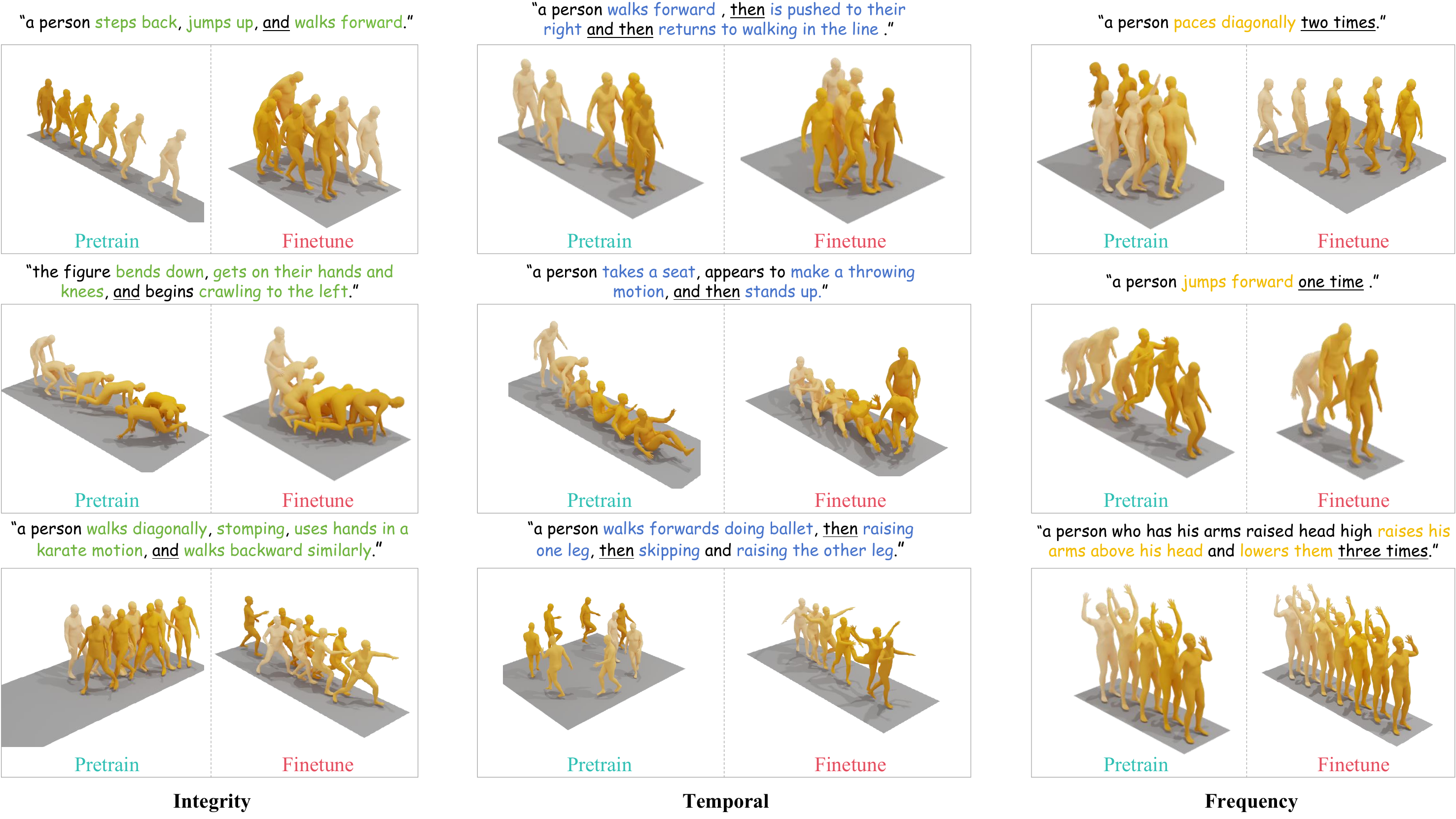}
    \vspace{-0.5cm}
    \caption{Generated qualitative samples comparison of pretrained model MotionGPT and finetuned model AToM.}
    \label{fig:qua}
    \vspace{-0.1cm}
\end{figure*}
\noindent\textbf{Quantitative Experiment} In our quantitative experiments across the temporal, frequency, and integrity tasks, AToM consistently outperforms MotionGPT \cite{jiang2023motiongpt} and InstructMotion \cite{sheng2024exploring} across most evaluation metrics, demonstrating superior text-motion alignment, motion quality, and generative realism. As shown in Table~\ref{tab:main_results}, AToM outperforms baselines in text-motion alignment, achieving lower MM Dist scores, higher top-1 and top-3 retrieval accuracy, and comparable top-2 accuracy, demonstrating its ability to generate semantically aligned motions. In terms of motion quality, AToM achieves a lower FID (0.613 vs. 0.655, a 6.4\% improvement), indicating more realistic motions. AToM balances alignment and diversity, with slight reductions in diversity and MModality attributed to fine-tuning focused on aligning with high-quality samples. In the general task, AToM excels with the lowest MM Dist (3.943), superior retrieval accuracy, and the best FID (0.177), showcasing its ability to generate realistic, well-aligned motions while maintaining consistency and robust generalization across sub-tasks and datasets. These improvements stem from our fine-tuning approach, which integrates AI feedback to better capture motion nuances valued by humans. By combining human-like feedback with efficient AI strategies, AToM surpasses traditional human-feedback-based models, offering a more effective and scalable solution.
\noindent\textbf{Qualitative Experiment} Figure~\ref{fig:qua} compares the MotionGPT \cite{jiang2023motiongpt} and AToM in terms of generation faithfulness for integrity, temporal and frequency tasks. In our qualitative analysis, discernible discrepancies are noted in the motion generation by the original model. For integrity task, in response to the three-event description ``\textit{a person steps back\textbf{,} jumps up, \textbf{and} walks forward.}", the generated motion included only two of the elements, but omitting specific motion event ``\textit{\textbf{jumps up}}". Similarly, for temporal task, the original model misrepresented the sequence of events or omitted action events. An example is the prompt ``\textit{a person walks forward, \textbf{then} is pushed to their right \textbf{and then} returns to walking in the line.}", where the generated motion incorrectly rendered the ``\textit{\textbf{is pushed to their right}}". For frequency task, the pretrained model generated motion ``\textit{a person jumps forward two times"}, which is inconsistent with the prompt ``\textit{\textbf{one time}}".
In contrast, AToM demonstrates enhanced performance, achieving more faithful generation on diverse motion events, complex temporal order and specific frequency shown in prompts.
\begin{figure}[h!]
    \centering
    \begin{minipage}{0.23\textwidth} 
        \vspace{0.01cm}
        \centering
        \includegraphics[width=\textwidth]{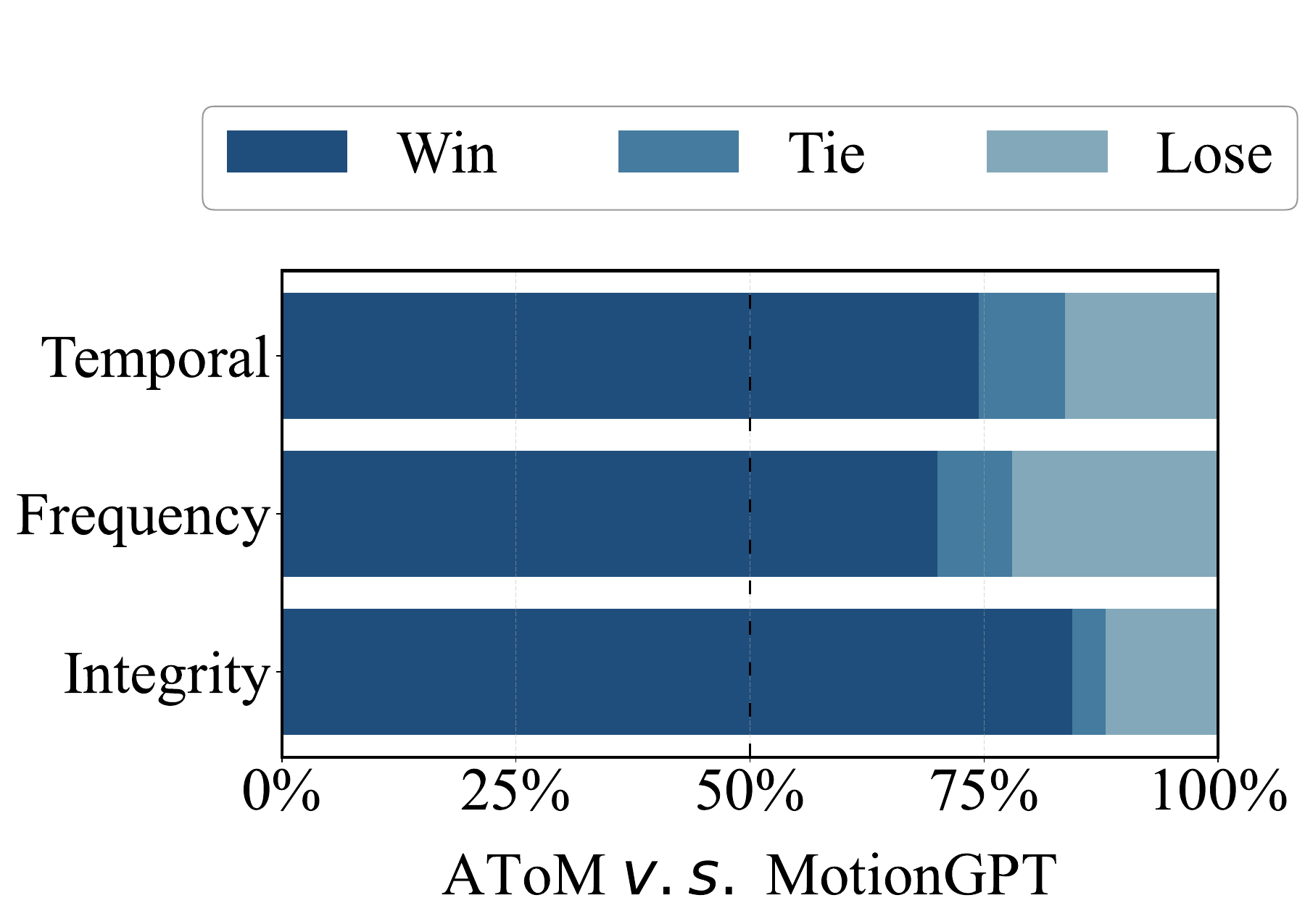}
        \vspace{0.01cm}
        \caption{Win rates of AToM fine-tuned compared to MotionGPT by human judgments in three tasks.}
        \label{fig:win}
    \end{minipage}
    \hspace{0.01\textwidth} 
    \begin{minipage}{0.22\textwidth} 
        \centering
        \includegraphics[width=\textwidth]{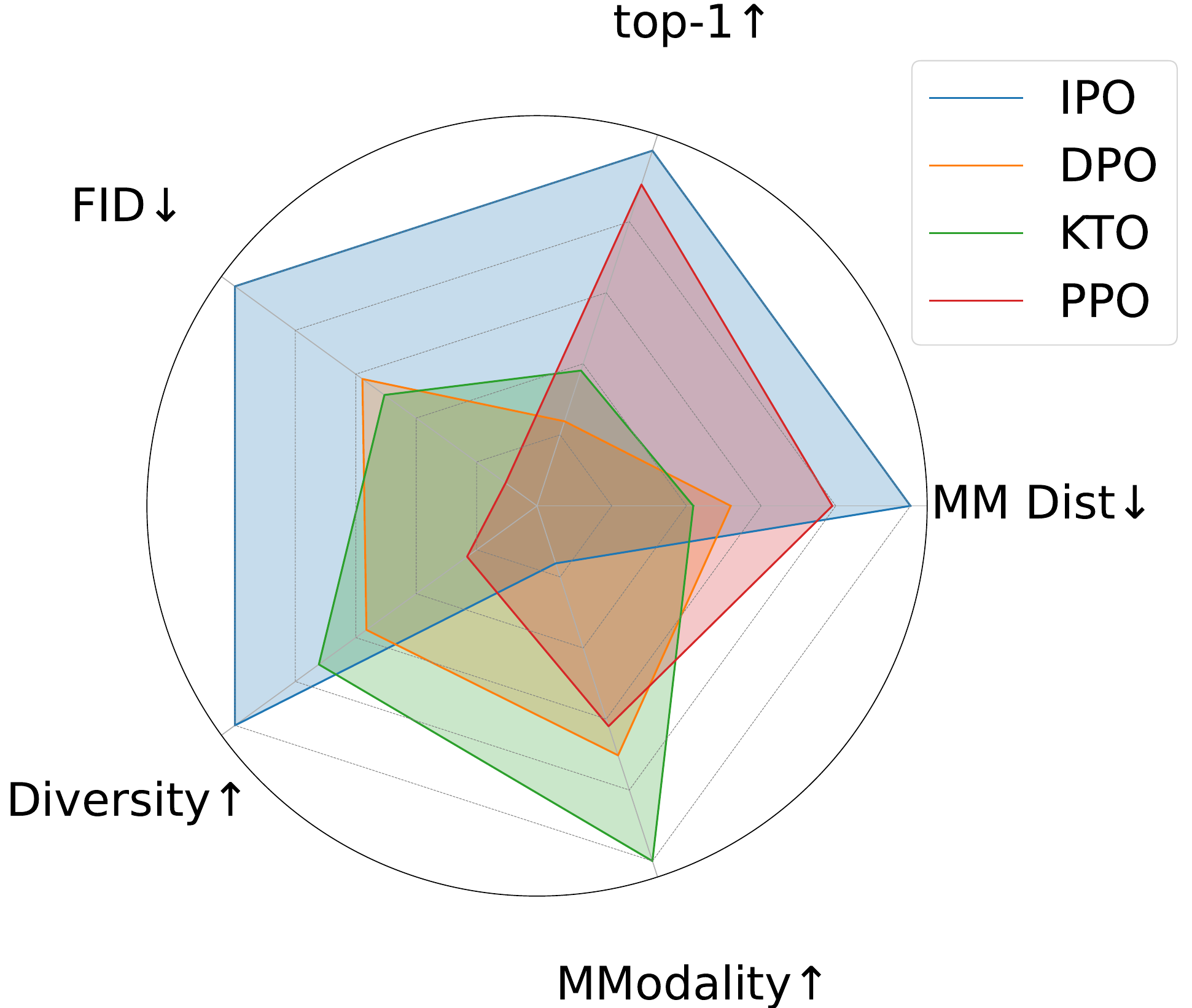}
        \caption{Performance distribution of different reinforcement learning strategies after generative model finetuning.}
        \label{fig:Radar}
    \end{minipage}
    \vspace{-0.5cm}
\end{figure}
\noindent\textbf{User Study} We conducted a human evaluation to compare the performance of AToM with the MotionGPT baseline. For each sub-task, motions were generated from randomly selected prompts in the filtered HumanML3D test set~\cite{Guo_2022_CVPR} mentioned in sec 4.1. Each of the 50 participants evaluated five randomly selected pairs of generated motions each across three criteria: frequency, integrity, and temporal alignment. For each pair, the participant chose the better motion or marked a tie if both were comparable. Figure \ref{fig:win} presents the average win and tie rates for AToM compared to MotionGPT. As shown, AToM outperforms MotionGPT across all sub-tasks, with win rates of 74.4\% for temporal, 70.0\% for frequency, and 84.4\% for integrity. This result reflects human evaluators' recognition of AToM's superior alignment and highlights its robustness across tasks.
\subsection{Ablation study}
\paragraph{Motion Injection Forms in Questioning}
To evaluate the effectiveness of different motion injection strategies in GPT-4V questioning, we experimented with three methods: (1) \textbf{Frame-by-Frame}: The generated motion sequences from MotionGPT \cite{jiang2023motiongpt} were rendered into video then sampled at 8-frame intervals, creating a sequence of images. (2) \textbf{Full-Image}: The sampled image sequence (at 8-frame intervals) was arranged into a composite image, with 5 frames per row. (3) \textbf{Trajectory-Image}: The motion data, sampled at 8-frame intervals from the motion sequence, were rendered by rigged cylinders to a single image. Injection demos are shown in the Appendix.
Table \ref{tab:merged_ablation_study} (a) shows that employing the Frame-by-Frame image sequence as the motion representation yielded the highest performance, with MM Dist (5.576), top-1 accuracy (0.199), and FID (0.613). This approach enhanced the fine-tuned model, leading to high-quality output and superior motion-text alignment. In contrast, Full-Image and Trajectory-Image yielded slightly lower performance, possibly due to less detailed frame-level information interpretable by GPT-4V.
\begin{table*}[htp!]
\centering
\caption{Ablation studies for motion injection methods, score filtering, and LoRA utilization on the test set.}
\begin{tabular}{l|ccc|cc|cc}
\hline
\textbf{Metric} & \multicolumn{3}{c|}{\textbf{(a) Motion Injection}} & \multicolumn{2}{c|}{\textbf{(b) Score Filtering}} & \multicolumn{2}{c}{\textbf{(c) LoRA Utilization}} \\
 & \textbf{F-b-F} & \textbf{F-I} & \textbf{T-I} & \textbf{w/ Filter} & \textbf{w/o Filter} & \textbf{w/ LoRA} & \textbf{w/o LoRA} \\
\hline
MM Dist $\downarrow$ & \cellcolor{gray!20}$\mathbf{5.576^{\pm .027}}$ & $5.657^{\pm .048}$ & $5.635^{\pm .036}$ & \cellcolor{gray!20}$\mathbf{5.576^{\pm .027}}$ & $5.640^{\pm .030}$ & \cellcolor{gray!20}$\mathbf{5.576^{\pm .027}}$ & $6.425^{\pm .066}$ \\
top-1 $\uparrow$ & \cellcolor{gray!20}$\mathbf{0.199^{\pm .005}}$ & $0.197^{\pm .008}$ & $0.191^{\pm .008}$ & \cellcolor{gray!20}$\mathbf{0.199^{\pm .005}}$ & $0.187^{\pm .005}$ & \cellcolor{gray!20}$\mathbf{0.199^{\pm .005}}$ & $0.128^{\pm .006}$ \\
top-2 $\uparrow$ & \cellcolor{gray!20}$\mathbf{0.322^{\pm .006}}$ & $0.316^{\pm .007}$ & $0.304^{\pm .007}$ & \cellcolor{gray!20}$\mathbf{0.322^{\pm .006}}$ & $0.305^{\pm .006}$ & \cellcolor{gray!20}$\mathbf{0.322^{\pm .006}}$ & $0.221^{\pm .007}$ \\
top-3 $\uparrow$ & \cellcolor{gray!20}$\mathbf{0.412^{\pm .008}}$ & $0.408^{\pm .005}$ & $0.399^{\pm .007}$ & \cellcolor{gray!20}$\mathbf{0.412^{\pm .008}}$ & $0.394^{\pm .008}$ & \cellcolor{gray!20}$\mathbf{0.412^{\pm .008}}$ & $0.296^{\pm .009}$ \\
FID $\downarrow$ & \cellcolor{gray!20}$\mathbf{0.613^{\pm .044}}$ & $0.803^{\pm .063}$ & $0.652^{\pm .042}$ & \cellcolor{gray!20}$\mathbf{0.613^{\pm .044}}$ & $0.693^{\pm .059}$ & \cellcolor{gray!20}$\mathbf{0.613^{\pm .044}}$ & $2.131^{\pm .144}$ \\
Diversity $\uparrow$ & \cellcolor{gray!20}$\mathbf{8.926^{\pm .086}}$ & $8.866^{\pm .103}$ & $8.841^{\pm .081}$ & \cellcolor{gray!20}$\mathbf{8.926^{\pm .086}}$ & $8.882^{\pm .085}$ & \cellcolor{gray!20}$\mathbf{8.926^{\pm .086}}$ & $8.582^{\pm .105}$ \\
MModality $\uparrow$ & \cellcolor{gray!20}$\mathbf{3.495^{\pm .147}}$ & $3.396^{\pm .128}$ & $3.675^{\pm .143}$ & \cellcolor{gray!20}$3.495^{\pm .147}$ & $\mathbf{3.564^{\pm .129}}$ & \cellcolor{gray!20}$3.495^{\pm .147}$ & $\mathbf{6.153^{\pm .182}}$ \\
\hline
\end{tabular}
\vspace{-0.4cm}
\label{tab:merged_ablation_study}
\end{table*}


\noindent\textbf{Score Filtering} As shown in Table \ref{tab:merged_ablation_study} (b), using score filtering in preference pairs construction, where only samples rated above three are considered positive-leads, results in clear performance improvements. With filtering, the model achieves an improved MM Dist, lowering the score from 5.640 to 5.576. The top-1 accuracy increases by 6.4\%, rising from 0.187 to 0.199, and the FID score decreases by 11.5\%, dropping from 0.693 to 0.613, all indicating better alignment and realism in generated motions. A larger faithfulness gap between positive and negative samples enhances the fine-tuning signal, enabling the model to better distinguish high-quality motion-text alignments.

\noindent\textbf{Finetune with LoRA}
In Table \ref{tab:merged_ablation_study} (c), incorporating LoRA\cite{hu2021lora} yields significant improvements compared to the model without it, though with a slight trade-off in multi-modality. Specifically, using LoRA\cite{hu2021lora} enhances the model’s ability to retrieve motion sequences accurately, reduces the MM Dist score from 6.425 to 5.576, and increases top-1 accuracy by 55.5\%, from 0.128 to 0.199. The top-2 and top-3 accuracies also improve significantly, with increases of 45.7\% and 39.2\%, respectively. The FID score also decreases by 71.2\%, from 2.131 to 0.613, indicating better alignment with real motion distributions. Diversity also increases, from 8.582 to 8.926, showing greater diversity in generated motions.

\noindent\textbf{RL Strategy}
Notably, Figure \ref{fig:Radar} (FID has been negatively treated) highlights that, among the four RL strategies, IPO shows the best overall performance. As the variant of DPO, IPO was specifically designed to mitigate overfitting\cite{azar2024general} associated with the Bradley-Terry(BT) model, resulting in better overall retrieval accuracy, alignment precision, and diversity compared to DPO, KTO, and PPO.

\begin{figure}[h!]
    \centering
    \vspace{-0.3cm}
    \includegraphics[width=0.48\textwidth]{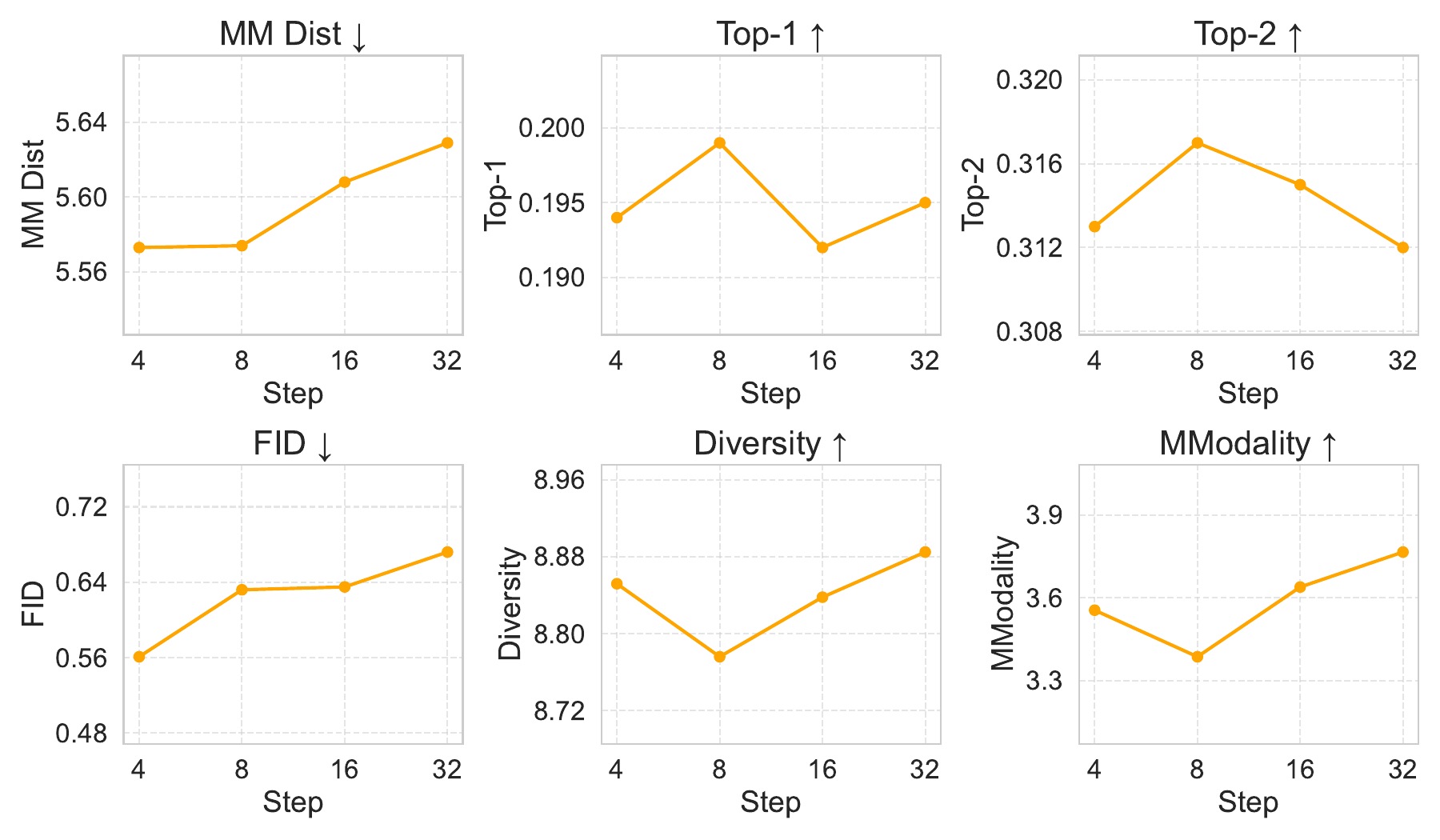}
    \vspace{-0.5cm}
    \caption{Impact of different frame sampling intervals on alignment and quality metrics.}
    \label{fig:step}
    \vspace{-0.6cm}
\end{figure}

\noindent\textbf{Sampling Steps of Motion Sequence}
Sampling frames at different intervals impacts performance, as illustrated in Figure \ref{fig:step}, with shorter intervals (e.g., 4 and 8) generally resulting in better match distances, higher top-1 accuracy, and lower FID, which indicate improved alignment and generation quality. We choose an interval of 8 frames as it provides strong retrieval precision, which is our primary focus.
\section{Conclusion}
In this paper, we introduce a novel framework, AToM, which utilizes GPT-4Vision feedback to enhance the alignment between text prompts and generated motions in text-to-motion models. AToM comprises three main stages: (1) generating diverse motions from constructed text prompts; (2) evaluating text-motion alignment using GPT-4Vision to construct the high-quality \texttt{MotionPrefer} dataset; and (3) fine-tuning the motion generator on \texttt{MotionPrefer}. Comprehensive quantitative and qualitative experiments demonstrate that AToM can effectively leverage feedback from Vision-Language Large Models, significantly improving text-motion alignment quality and paving the way for advancements in motion synthesis from textual prompts.
{
    \clearpage
    \newpage
    \small
    \bibliographystyle{ieeenat_fullname}
    \bibliography{main}
}
\clearpage
\setcounter{page}{1}
\maketitlesupplementary
\section{Additional Results}
\subsection{More Qualitative Results}
\begin{figure*}[h!]
    \vspace{-0.4cm}
    \centering
    \includegraphics[width=\textwidth]{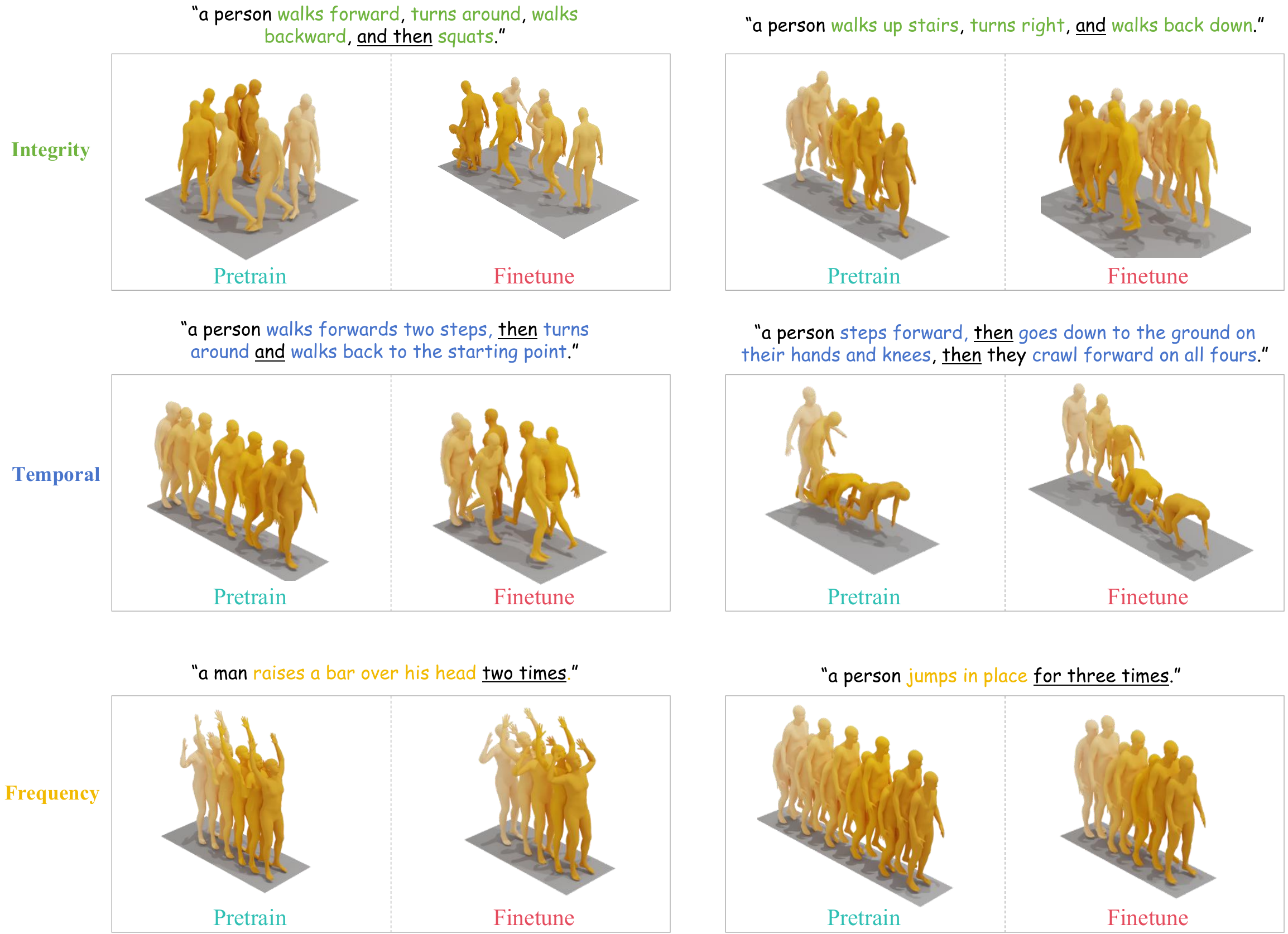}
    \vspace{-0.5cm}
    \caption{Generated qualitative samples comparison of pretrained model MotionGPT and finetuned model AToM.}
    \label{fig:sup_qua}
    \vspace{-0.1cm}
\end{figure*}

Figure \ref{fig:sup_qua} presents additional qualitative comparisons between AToM and the baseline models, highlighting AToM's superior performance.

\subsection{Number of Iterations for Fine-tuning}
\begin{figure}[h!]
    \centering
    \includegraphics[width=0.48\textwidth]{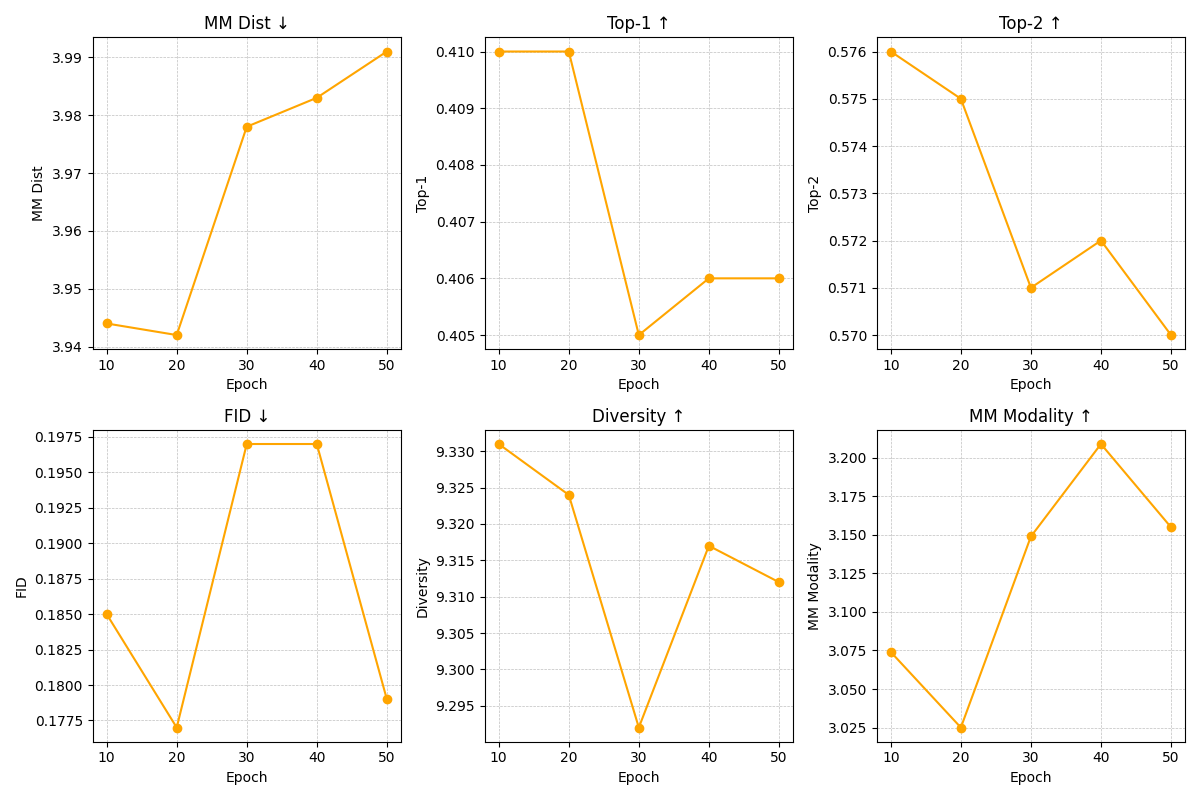}
    \caption{Impact of different epoch numbers on alignment and quality metrics.}
    \label{fig:supp_epoch}
    \vspace{-0.3cm}
\end{figure}

We increased the number of iterations for fine-tuning, with the results presented in Figure~\ref{fig:supp_epoch}. In the general task, which includes data from three sub-tasks, increasing fine-tuning iterations has a mixed impact on the evaluated metrics. During the early stages, performance improves across most metrics, including reductions in MM Dist and FID, along with higher Top-1/Top-2 accuracy and Diversity, reflecting enhanced sample quality, diversity, and text-motion alignment. However, exceeding 30 iterations tends to lead to overfitting, resulting in degraded generalization, as seen in increased MM Dist and fluctuations in Top-1/Top-2 accuracy. This suggests that prolonged fine-tuning reduces the model's ability to generalize across distributions and may lead to mode collapse, thereby negatively impacting diversity and FID. Optimal performance is observed between 20-30 iterations, where the balance between quality and generalization is most effectively maintained.

\subsection{Hyper-parameter $\beta$ of IPO}
\begin{figure}[h!]
    \centering
    \includegraphics[width=0.48\textwidth]{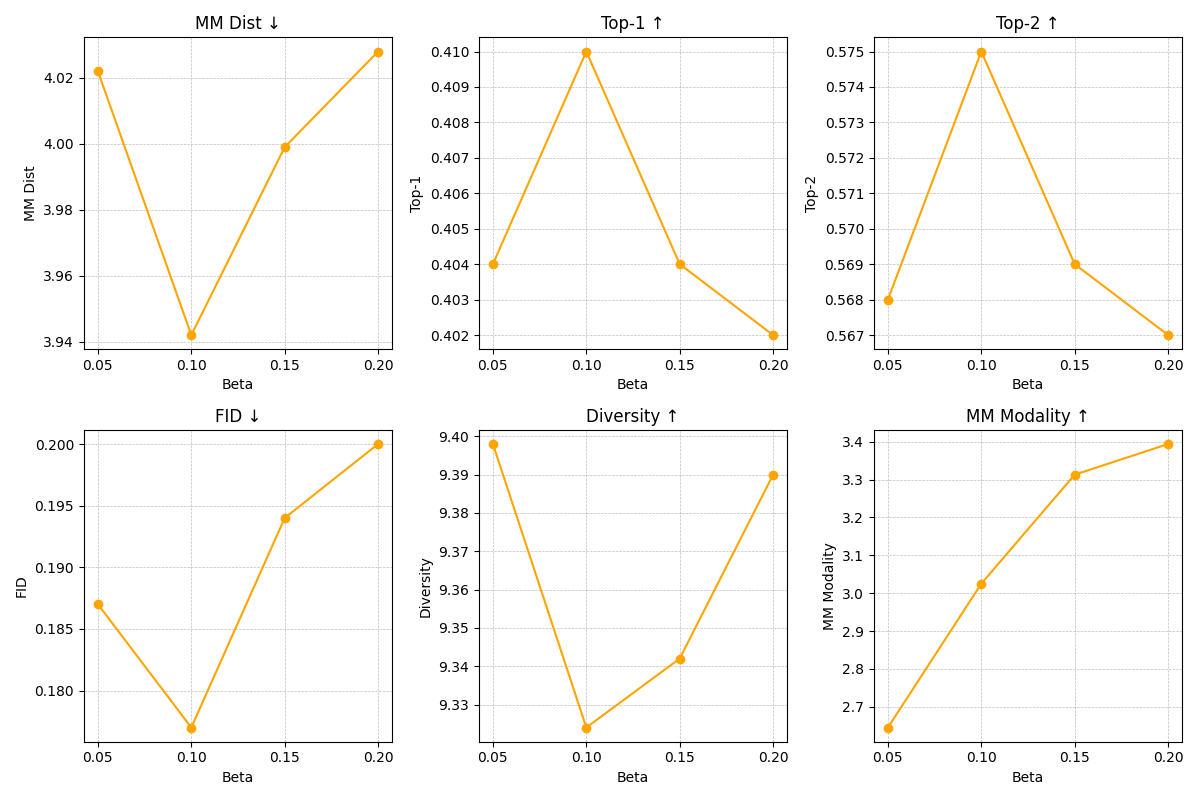}
    \caption{Impact of $\beta$ on alignment and quality metrics.}
    \label{fig:beta}
    \vspace{-0.3cm}
\end{figure}

The effect of the IPO hyper-parameter \(\beta\) on alignment and quality metrics is illustrated in Figure~\ref{fig:beta}. The optimal performance across most metrics is achieved at \(\beta = 0.10\), where MM Dist and FID are minimized, and Top-1/Top-2 accuracy and Diversity reach their maximum values, indicating enhanced alignment, sample quality, and diversity. However, increasing \(\beta\) beyond 0.10 leads to increased MM Dist and FID, likely caused by an overemphasis on alignment objectives. In contrast, smaller \(\beta\) values (e.g., \(\beta = 0.05\)) fail to adequately align the model, resulting in suboptimal performance. These findings highlight the inherent trade-offs between alignment, quality, and diversity, underscoring the importance of setting \(\beta = 0.10\) to effectively balance these competing objectives.

\subsection{Preference Accuracy Comparison: GPT-4V vs. Contrastive Encoders on Human Preference Datasets}

In the introduction, we noted that prior works, such as Mao \etal~\cite{mao2024learning}, have utilized contrastive pre-trained text and motion encoders from Guo \etal~\cite{Guo_2022_CVPR} to construct reward models. Following this approach, we conducted an evaluation to compare the alignment accuracy of the contrastive encoders and our proposed method against human preferences.

To assess the ability of GPT-4V and the contrastive encoders to align with human preferences, we evaluated their alignment accuracy using the human preference dataset provided by InstructMotion \cite{sheng2024exploring}. For each pair in the dataset (excluding those marked as ``skipped"), GPT-4V was prompted to evaluate the motions and determine which one performed better, providing its preference directly. For the contrastive encoders, we encoded both the motion and text features separately and calculated the Euclidean distance between the two. The motion in the pair with the smaller distance to the text feature was considered the preferred sample. This setup allowed us to systematically compare the two approaches' alignment performance with human annotations.

As shown in Table \ref{tab:alignment_quality}, among the total of 2216 pairs, GPT-4V achieved an alignment accuracy of $69.77\%$ , with 1546 aligned pairs. In contrast, the contrastive encoders exhibited a lower alignment accuracy of $66.11\%$, with 1465 aligned pairs out of the same total. These results highlight the superior capability of GPT-4V in capturing human preferences compared to the contrastive encoders, underscoring its potential for more effective human-centric applications.

\begin{table}[h!]
\centering
\resizebox{\columnwidth}{!}{
\begin{tabular}{lccc}
\toprule
\textbf{Model} & \textbf{Aligned Pairs} & \textbf{Total Pairs} & \textbf{Accuracy (\%)} \\
\midrule
GPT-4V & 1546 & 2216 & 69.77 \\
CE-based   & 1465 & 2216 & 66.11 \\
\bottomrule
\end{tabular}
}
\caption{Alignment quality comparison between GPT-4V and contrastive encoder-based methods (denoted as CE-based)}
\label{tab:alignment_quality}
\end{table}

\subsection{GPT-4V Finetuned v.s. Constrastive Encoders Finetuned}

To better demonstrate that our method outperforms approaches leveraging contrastive pre-trained encoders, we utilized these encoders to label motions generated in our temporal sub-task. The labeled preference data was then used to fine-tune MotionGPT, and a comparative analysis was conducted against AToM. As presented in Table \ref{tab:ours_vs_cmlp}, AToM consistently outperforms the approach based on contrastive encoders across key metrics, including MM Dist, R-precision, FID, and MultiModality, demonstrating its superior capability in alignment quality and generation variety. Although AToM exhibits slightly lower performance in the Diversity metric, its overall advantage across other critical metrics underscores its effectiveness and robustness compared to the contrastive encoder-based approach.

\begin{table*}[ht]
\centering
\caption{Comparison of methods AToM (ours) and contrastive encoder-based method. }
\begin{tabular}{@{}lccccccc@{}}
\toprule
\textbf{Method} & \textbf{MM Dist$\downarrow$} & \textbf{Top-1$\uparrow$} & \textbf{Top-2$\uparrow$} & \textbf{Top-3$\uparrow$} & \textbf{FID$\downarrow$} & \textbf{Diversity$\uparrow$} & \textbf{MultiModality$\uparrow$} \\ 
\midrule
AToM (ours) & $5.576^{\pm .027}$ & $0.232^{\pm .004}$ & $0.354^{\pm .006}$ & $0.432^{\pm .005}$ & $0.539^{\pm .031}$ & $8.793^{\pm .074}$ & $3.736^{\pm .147}$ \\
CE-based & $5.664^{\pm .028}$ & $0.188^{\pm .005}$ & $0.307^{\pm .006}$ & $0.392^{\pm .005}$ & $0.556^{\pm .039}$ & $8.843^{\pm .096}$ & $3.724^{\pm .131}$ \\
\bottomrule
\end{tabular}
\label{tab:ours_vs_cmlp}
\end{table*}

\subsection{Influence of Preference Dataset Volume on Model Performance}
\begin{figure}[h!]
    \centering
    \includegraphics[width=0.48\textwidth]{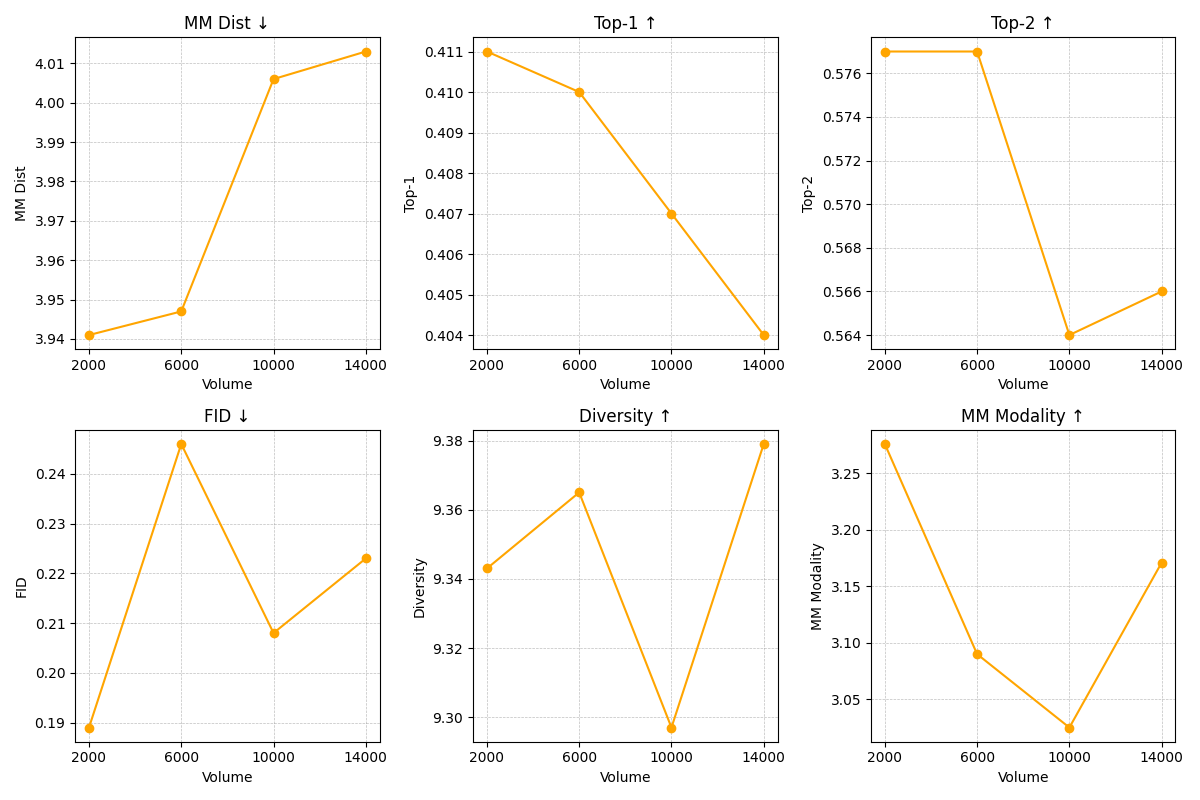}
    \caption{Impact of preference pair quantity on alignment and quality metrics.}
    \label{fig:supp_volume}
    \vspace{-0.3cm}
\end{figure}

The impact of preference pair quantity on alignment and quality metrics is depicted in Figure~\ref{fig:supp_volume}. At lower volumes (e.g., 2000 pairs), metrics such as MM Dist and FID are minimized, indicating better alignment and motion quality, while Diversity and MM Modality are relatively high, suggesting balanced performance. However, as the volume increases, MM Dist and FID worsen, and Top-1/Top-2 accuracy decreases significantly, likely due to over-fitting to a larger but potentially noisy set of preferences, which degrades generalization. The Diversity and MM Modality metrics exhibit fluctuations, with notable drops at intermediate volumes (e.g., 10,000 pairs) and partial recovery at higher volumes (14,000 pairs). These observations highlight the trade-off between data volume and model performance, where excessively large preference datasets may introduce noise, reducing alignment and diversity, and emphasizing the need for careful curation and optimal dataset sizing.

\section{Details of \texttt{MotionPrefer} Construction}
\subsection{GPT-4 Instruction for Prompt Construction}
We instruct GPT-4 to generate motion-event-based prompt. The designed instruction for three tasks are as follows:

\begin{table}[h]
\centering
\small
\begin{tcolorbox}[]

Given a label group, $\mathrm{X_{task}}$, the 2-5 labels in it will be described into a motion event group:\\ $\{ \texttt{Event}_{1}, ..., \texttt{Event}_{5} \}$.\\ Then, please join the motion event group with conjunction to form a motion prompt, defined as\\ ``$\texttt{Event}_{1}, \texttt{Conjunction}_{1}, \texttt{Event}_{2}, \texttt{Conjunction}_{2},\\ ..., \texttt{Conjunction}_{4}, \texttt{Event}_{5}$". 
\\
\\
Conjunction list: $\mathrm{X_{conj}}$
\\
\\
Please ensure to randomly select conjunctions from the list and avoid relying on a single conjunction. Please try to generate complete prompts that match human language expressions as much as possible.
\end{tcolorbox}
\caption{GPT instruction for prompt construction in integrity task.}
\vspace{-0.4cm}
\label{tab:gpt_construction}
\end{table}

The distribution of prompts with varying numbers of motion events is shown in the Table \ref{tab:motion_event_distribution}.
\begin{table}[h]
\centering
\begin{tabular}{c|c}
\hline
\textbf{Motion events} & \textbf{Prompt} \\ \hline
2 & 250 \\ \hline
3 & 1500 \\ \hline
4 & 500 \\ \hline
5 & 250 \\ \hline
\end{tabular}
\caption{Motion event number distribution.}
\label{tab:motion_event_distribution}
\end{table}

\begin{table}[h]
\centering
\small
\begin{tcolorbox}[]

Given a label group, $\mathrm{X_{task}}$, the 3 labels in it will be described into a motion event group:\\ $\{ \texttt{Event}_{1}, \texttt{Event}_{2}, \texttt{Event}_{3} \}$.\\ Then, please join the motion event group with conjunction to form a motion prompt, defined as\\ ``$\texttt{Event}_{1}, \texttt{Conjunction}_{1}, \texttt{Event}_{2}, \texttt{Conjunction}_{2}, \\
\texttt{Event}_{3}$". 
\\
\\
Conjunction list: $\mathrm{X_{conj}}$
\\
\\
Please ensure ...
\end{tcolorbox}
\caption{\small{GPT instruction for prompt construction in temporal task.}}
\label{tab:gpt_construction}
\end{table}

\begin{table}[h]
\centering
\small
\begin{tcolorbox}[]

Given a label group, $\mathrm{X_{task}}$, the single label in it will be described into a motion event: $\{ \texttt{Event}_{1} \}$.\\ Then, please join the motion event with frequency to form a motion prompt, defined as ``$\texttt{Event}_{1}, \texttt{Frequency}_{1}$". 
\\
\\
Frequency list: $\mathrm{X_{freq}}$
\\
\\
Please ensure ...
\end{tcolorbox}
\caption{GPT instruction for prompt construction in frequency task.}
\vspace{-0.4cm}
\label{tab:gpt_construction}
\end{table}

\subsection{GPT Instruction for Scoring}
In this section, we outline the GPT-based instructions for scoring, providing a comprehensive framework for evaluating alignment between motion and description effectively.
\begin{table}[h!]
\centering
\vspace{0cm}
\small
\begin{tcolorbox}[]
Please evaluate the alignment between a given generative motion clip and the corresponding text description (``Input"). The description $\mathrm{T}$, consists of 2-5 motions, and the motion clip is represented as a sequence of frames $\mathrm{M}$.
\\
\\
For example:
    
    $\texttt{T}$= ``A man walks forward, walks backward, and squats" describes three motions."
\\
\\
Scoring Criteria:

    5: All described motions appear in the frames.
    
    0: Some motions are missing or incomplete in the frames.
\\
\\
Output Format:

    Rating: [0 or 5]
    
    Rationale: [A brief explanation for the rating, no more than 20 words]
\end{tcolorbox}
\caption{GPT annotation instruction for integrity task.}
\label{tab:gpt}
\end{table}

\begin{table}[h!]
\centering
\small
\begin{tcolorbox}[]
Please evaluate the alignment between a given generative motion clip and the corresponding text description ("Input"). The description $\mathrm{T}$, describes 3 motions in temporal order, and the motion clip is represented as a sequence of frames $\mathrm{M}$.
\\
\\
For example:
    
    $\texttt{T}$= ``A person walks forward , then is pushed to their right and then returns to walking in the line."
\\
\\
Scoring Criteria:

    5: Three motions appear in correct order.
    
    4: Three motions appear in wrong order.
    
    3: One motion is missing.
    
    2: Two motions are missing.
    
    1: All three motions are missing.
\\
\\
Output Format:

    Rating: [1 to 5]
    
    Rationale: [A brief explanation for the rating, no more than 20 words]
\end{tcolorbox}
\caption{GPT annotation instruction for temporal task.}
\label{tab:gpt}
\end{table}

\begin{table}[h!]
\centering
\small
\begin{tcolorbox}[]
Please evaluate the alignment between a given generative motion clip and the corresponding text description (``Input"). The description $\mathrm{T}$, describes a repeated motion for several times, and the motion clip is represented as a sequence of frames $\mathrm{M}$.
\\
\\
For example:
    
    $\texttt{T}$= ``a person jumps forward three times.“
\\
\\
Scoring Criteria:

    3: The motion is correct and the frequency is accurate.
    
    2: The motion is present but the frequency is incorrect.
    
    1: The motion is incorrect, regardless of the frequency.
\\
\\
Output Format:

    Rating: [1 to 3]
    
    Rationale: [A brief explanation for the rating, no more than 20 words]
\end{tcolorbox}
\caption{GPT annotation instruction for frequency task.}
\label{tab:gpt}
\end{table}

\subsection{Motion Injection Forms in Questioning}
\begin{figure}[h!]
    \centering
    \vspace{-0.5cm}
    \includegraphics[width=0.36\textwidth]{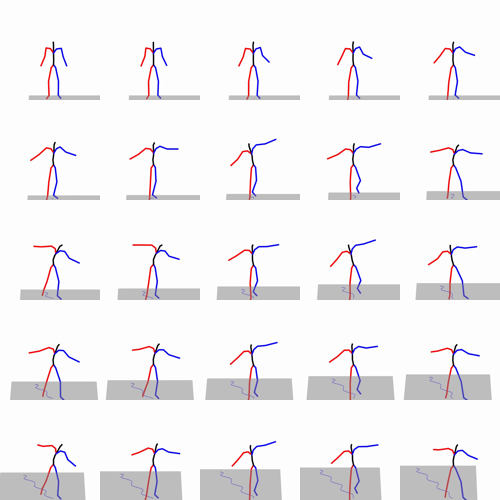}
    \caption{\textbf{Full-Image Example}}
    \label{fig:Full-Image}
\end{figure}

\begin{figure}[h!]
    \centering
    \includegraphics[width=0.3\textwidth]{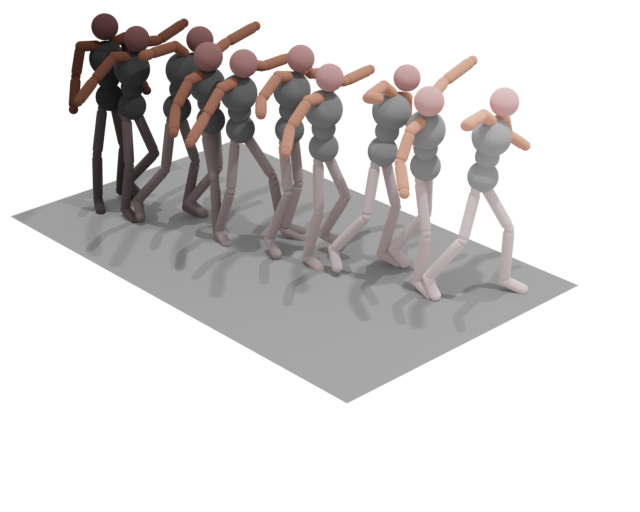}
    \vspace{-1cm}
    \caption{\textbf{Trajectory-Image Example}}
    \label{fig:Trajectory}
    \vspace{0.5cm}
\end{figure}

\begin{figure*}[t!]
    \vspace{-2cm}
    \centering
    \includegraphics[width=\textwidth]{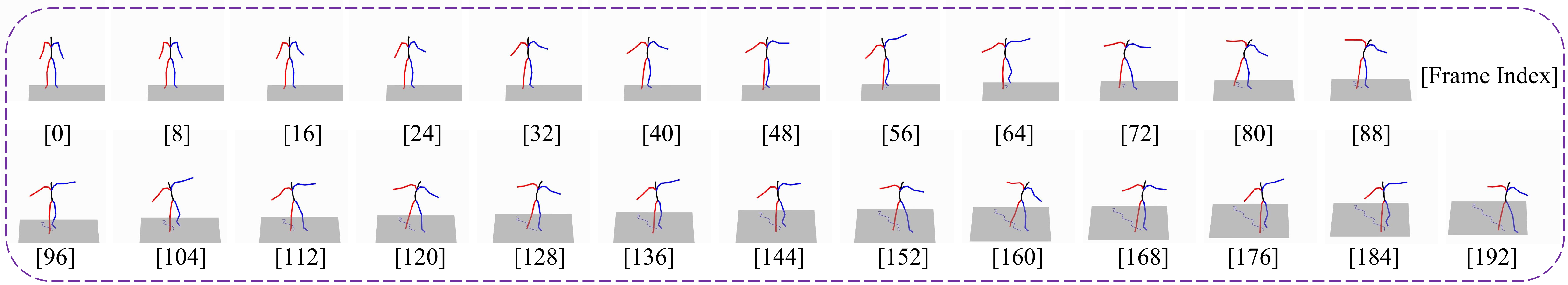}
    \caption{\textbf{Frame-by-Frame Example}}
    \label{fig:Frame-by-Frame}
    \vspace{-4cm}
\end{figure*}

There are demonstrations of three different forms of motion injection in questioning, as illustrated in Figure \ref{fig:Full-Image} to \ref{fig:Frame-by-Frame}.

\section{Human Evaluation}
\begin{figure*}[]
    \centering
    \includegraphics[width=0.8\textwidth]{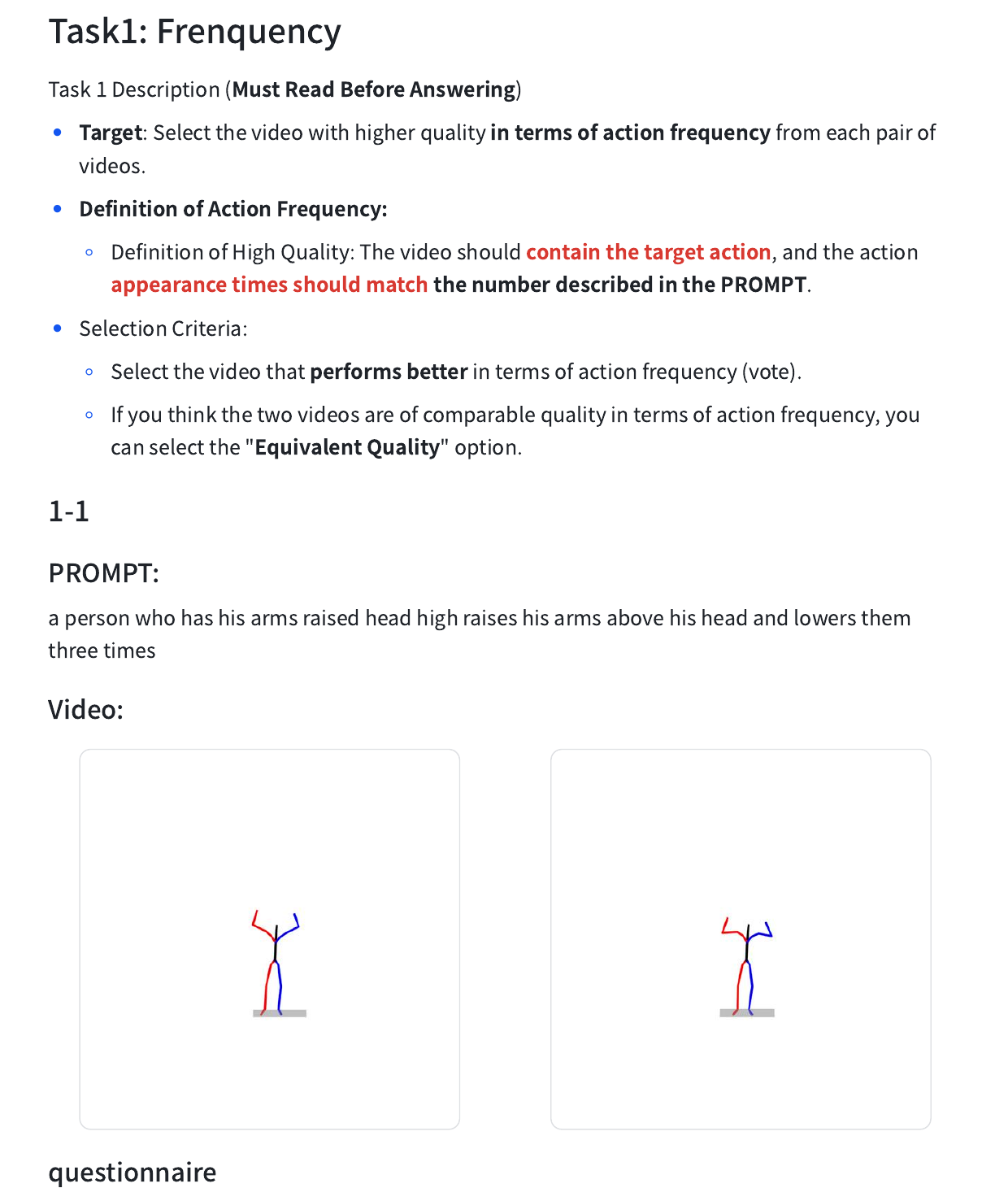}
    \caption{\textbf{User study example of frequency task}}
    \label{fig:user1}
    \vspace{-2cm}
\end{figure*}

We present an example of the user study for the frequency task in Figure~\ref{fig:user1}.

\end{document}